\def\year{2021}\relax
\documentclass[letterpaper]{article} 
\usepackage{aaai21}  
\usepackage{times}  
\usepackage{helvet} 
\usepackage{courier}  
\usepackage[hyphens]{url}  
\usepackage{graphicx} 
\usepackage{natbib}  
\usepackage{caption} 
\usepackage{times}
\usepackage{epsfig}
\usepackage{graphicx}
\usepackage{amsmath}
\usepackage{amssymb}
\usepackage{multirow}
\usepackage{tabularx}
\usepackage{booktabs}
\usepackage{mathtools}
\usepackage{mathtools}
\usepackage{enumerate}
\usepackage{xcolor}
\usepackage{booktabs}
\usepackage{multirow}
\usepackage{color, colortbl}
\usepackage{bigstrut}
\usepackage{amssymb}
\usepackage{floatrow}
\usepackage{caption}
\usepackage{arydshln}
\usepackage{subcaption}
\definecolor{myred}{rgb}{ .753,  0,  0}
\definecolor{myblue}{rgb}{0,  0,  1}

\title{Interpreting Deep Neural Networks with Relative Sectional Propagation\\
by Analyzing Comparative Gradients and Hostile Activations}
\author{
Woo-Jeoung Nam,\textsuperscript{\rm 1}
Jaesik Choi,\textsuperscript{\rm 3}
Seong-Whan Lee\textsuperscript{\rm 1,2}\thanks{Corresponding author: Seong-Whan Lee}
}
\affiliations {
\\\textsuperscript{\rm 1}Department of Computer and Radio Communications Engineering, Korea University, Seoul, Republic of Korea
\\\textsuperscript{\rm 2}Department of Artificial Intelligence, Korea University, Seoul, Republic of Korea
\\\textsuperscript{\rm 3}Graduate School of Artificial Intelligence, KAIST, Daejeon, Republic of Korea\\
}
\begin{document}
\maketitle

\begin{abstract}
The clear transparency of Deep Neural Networks (DNNs) is hampered by complex internal structures and nonlinear transformations along deep hierarchies. In this paper, we propose a new attribution method, Relative Sectional Propagation (RSP), for fully decomposing the output predictions with the characteristics of class-discriminative attributions and clear objectness. We carefully revisit some shortcomings of backpropagation-based attribution methods, which are trade-off relations in decomposing DNNs. We define hostile factor as an element that interferes with finding the attributions of the target and propagate it in a distinguishable way to overcome the non-suppressed nature of activated neurons. As a result, it is possible to assign the bi-polar relevance scores of the target (positive) and hostile (negative) attributions while maintaining each attribution aligned with the importance. We also present the purging techniques to prevent the decrement of the gap between the relevance scores of the target and hostile attributions during backward propagation by eliminating the conflicting units to channel attribution map. Therefore, our method makes it possible to decompose the predictions of DNNs with clearer class-discriminativeness and detailed elucidations of activation neurons compared to the conventional attribution methods. In a verified experimental environment, we report the results of the assessments: (i) Pointing Game, (ii) mIoU, and (iii) Model Sensitivity with PASCAL VOC 2007, MS COCO 2014, and ImageNet datasets. The results demonstrate that our method outperforms existing backward decomposition methods, including distinctive and intuitive visualizations.
\end{abstract}

\section{Introduction}
As Deep Neural Networks (DNNs) have shown a remarkable performance in various fields, many studies have attempted to resolve the basis of network predictions. However, it still lacks clear transparency about the myriad of components and the complex inner structure of DNNs. The problem of attribution, also called relevance, seeks the most relevant factors with respect to the predictions of DNNs and characterizes them as a supporting basis for the decision.
\begin{figure}
  \centering
  \includegraphics[width=1\linewidth]{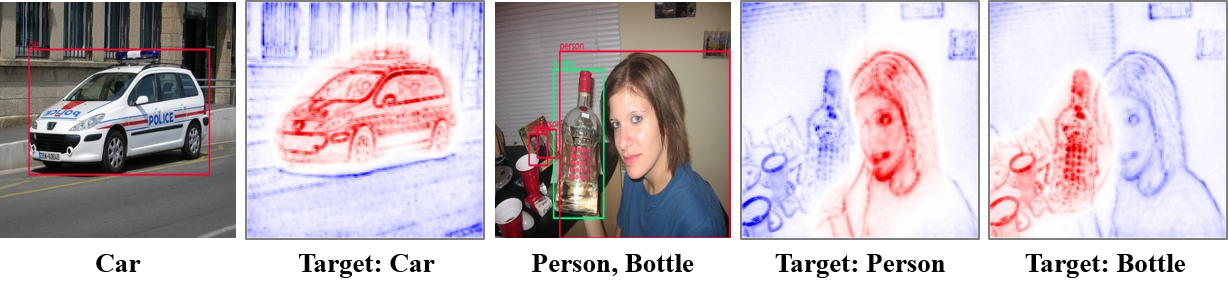}
    \caption{Relative Sectional Propagation (RSP) aims to fully decompose the network predictions with taking advantages of i) strong objectness, ii) class-discriminativeness, and iii) detailed descriptions of neuron activations.}
  \label{fig:intro}
\end{figure}
Grad-CAM \cite{selvaraju2017grad} is the most popular and widely used method in the field of weakly supervised segmentation and detection due to the easily applicable property and its high performance for localizing the primary objects. Despite these advantages, there is a limitation that the feature extraction stage of DNNs cannot be decomposed \cite{rebuffi2020there}. Also, detailed information about the neuron activation is lost due to the interpolation of the class activation map.
To fully interpret the network, many studies based on the modified backpropagation algorithm ~\cite{bach2015pixel,kindermans2017patternnet, montavon2017explaining, zhang2018top, nam2019relative} attempted to identify the significant parts in the input image with each of their own perspectives by decomposing the output predictions in a backward manner. By visualizing attributions as saliency maps, the important objects are highlighted as the basis for the predictions. Despite many studies on such methods, it is challenging to address the class-discriminativeness, incorrect distribution of attributions that are derived in unrelated objects, and model sensitivity. Furthermore, the different mechanisms among various types of DNNs and the variations of feature attention according to the layer depth make the interpretation of the network more difficult. 

The class discriminative issue was addressed in recent papers, and the contrastive perspective was presented as a countermeasure \cite{zhang2018top, gu2018understanding}. The idea of contrasting is erasing the duplicated attributions among classes by backpropagating twice from the target and all other classes. By efficiently removing the duplicate relevance of the activated neurons, it is possible to obtain the attributions of the target class within the object area. However, unreasonable positive or negative attributions in irrelevant regions, such as background and watermark, are easily found in the results of this concept. 

In this paper, we propose an attribution method, Relative Sectional Propagation (RSP), by analyzing the relative gradient activation maps between the target and hostile classes and propagating corresponding relevance according to the sectional influence of individual neurons. We carefully investigate the reasons for the non-suppressed characteristics of neuron activations according to the different classes and address these issues by assigning the bi-polar relevance scores: from highly related with target to highly relevant to hostile classes. Inspired by the traits of \textbf{winner always wins} ~\cite{zhang2018top} among activated neurons, the main idea is to separately compute the relative gradient activation maps, which contain the neuron importance, and purge the conflicting units to the channel attributions along the channel-axis, thereby preventing the decrement of the gap between bi-polar relevance scores. Our method preserves the conservation rule ~\cite{bach2015pixel} to prevent degeneracy problem, and allocates the relevance scores aligned with contributions.

Fig.~\ref{fig:intro} illustrates the samples that summarize the advantages of RSP. The attributions are fully decomposed from the output to input with the characteristics of i) the detailed visualizations of neuron activations, ii) strong objectness to output predictions, and iii) discriminativeness among classes with bi-polar relevance scores (positive and negative). As these characteristics are the trade-off relations in previous attribution literature, we mainly focus on overcoming the limitations during decomposition. 
The main contributions of this work are as follows:
\begin{itemize}
\item We propose a new method for decomposing the output predictions with the relative gradient activation maps and backward sectional propagation according to the individual influences of neurons. By hostilely changing the priority of attributions corresponding to the non-target, it is possible to properly distribute the bi-polar relevance scores between the predicted classes while maintaining the irrelevant attributions as negative.

\item 
We carefully address the phenomenon of non-suppressed characteristics of activated neurons and the contrary influences of the conflicting units to the channel attribution map, both of which prevent the attributions from being distinctive. We present a purging process to account for these neurons and to sustain the gap between positive and negative attributions.
\item 
For evaluation, we apply Pointing Game \cite{lapuschkin2016analyzing}, sanity check with Model Sensitivity \cite{adebayo2018sanity}, and mIOU to assess the quality of attributions. We report the performance in two cases of model decision (either only correct or all labels) to confirm the efficacy of interpreting the models. The evaluation demonstrates that our method outperforms other backpropagation-based attribution methods in circumstances of complete decomposition, including the advantages of strong objectness, class-discriminativeness, and detailed descriptions of activations.

\end{itemize}

\section{Related Work}
As DNNs are applied to a variety of traditional computer vision methodologies \cite{roh2007accurate,roh2010view,yang2007reconstruction,bulthoff2003biologically}, there are many attempts to improve the transparency issues of DNNs. As the manner of interpreting a DNN model itself, intermediate features are visualized by maximizing the activated neurons in intermediate layers ~\cite{erhan2009visualizing} or generating saliency maps \cite{simonyan2013deep,zeiler2014visualizing,mahendran2016visualizing,zhou2016learning,dabkowski2017real,zhou2018interpreting}. ~\cite{ribeiro2016should} proposed LIME, which explains the black-box models, by locally approximating them as simpler interpretable models.

A perturbation-based approach directly analyzes the variations of decision when distorting the input of the network. ~\cite{zeiler2014visualizing, petsiuk2018rise} investigate the variations of the output as applying occlusions to images with specified patterns. ~\cite{fong2019understanding} introduced the concept of extremal perturbation to understand network behavior with theoretically based masking.

From the viewpoint of decomposing the network decision, ~\cite{bach2015pixel} proposed several kinds of Layer-wise Relevance Propagation (LRP) rules with the concept of relevance and conservation. As a theoretical foundation, ~\cite{montavon2017explaining} proposed Deep Taylor decomposition by utilizing Taylor expansion among neurons of intermediate layers. \cite{selvaraju2017grad} proposed Grad-CAM to generate class-discriminative activation maps by computing gradients with respect to the last convolutional units of the feature extraction stage. Guided BackProp~\cite{springenberg2014striving} is based on gradient backpropagation that considers only positive values. Integrated Gradients \cite{sundararajan2017axiomatic} addressed the gradient saturation problem by computing the average partial derivatives of the output. DeepLIFT \cite{shrikumar2017learning} decomposes the differences in relevance scores between the activation and its reference. \cite{ancona2018towards} approached with the theoretical perspective to attributions and formally proved the conditions of equivalence of the previous methods. \cite{lundberg2017unified} unifies some explaining methods and approximate with the shapley value. ~\cite{zhang2018top} proposed Excitation Backprop (EB) by modeling the probabilistic winner-take-all process and addressed the class-discriminative issue with contrastive top-down attention. ~\cite{nam2019relative} pointed out the overlapping phenomenon of positive and negative relevance and utilized the influential perspective to separate relevant and irrelevant attributions. \cite{lapuschkin-ncomm19} discussed the spurious correlations among the objects in the input, (such as tags in pictures) and present the necessity of comprehending the network decision to unmask ``Clever Hans'' phenomenon.

We mainly focus on attribution methods based on backpropagation. Although there are many studies related to this, the interpretation of DNNs remains a trade-off between the objectives of each attribution method. Therefore, our method aims to overcome the main issues: class-discriminativeness, the details of neuron activations with full decomposition, and objectness which can separate the main objects and background.
\begin{figure}[!t]
  \centering
  \includegraphics[width=1\linewidth]{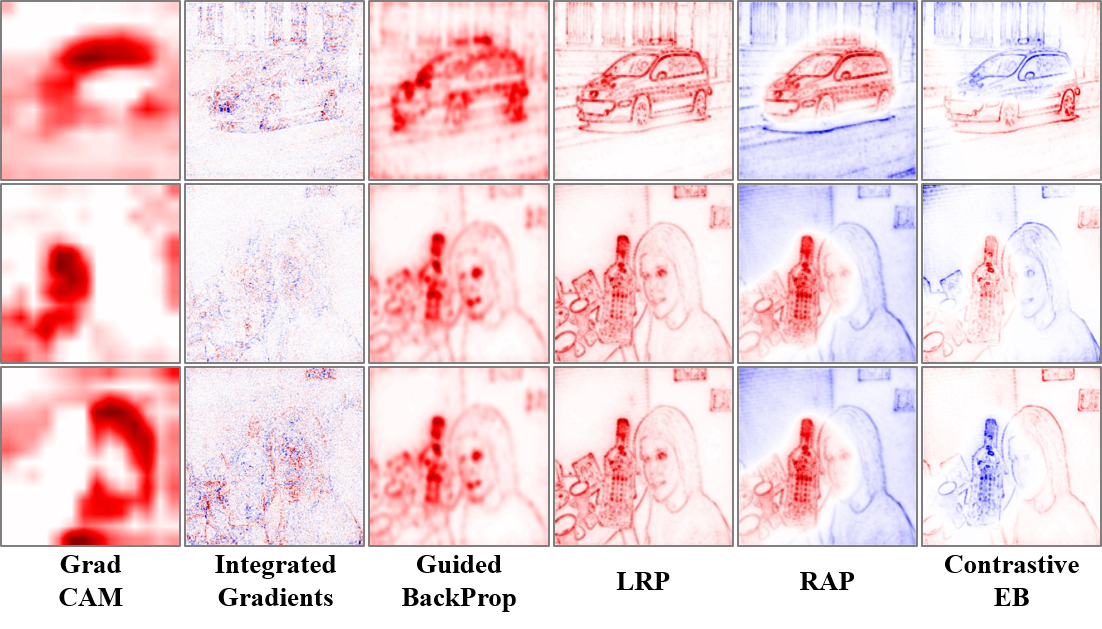}
    \caption{The shortcomings of some attribution methods. Results of Grad-CAM and Integrated Gradients are lack of detail. Gui, LRP, and RAP are not class discriminative. Some attributions of CEB are distributed in unrelated parts.}
  \label{fig:2}
\end{figure}

\section{Revisiting Attribution Methods}
In a multi-classification task, it is clear that objects in the input are not learned antagonistically during the training procedure, because most of the networks consider the correct predictions among output logits simultaneously, not competitively. When we investigate the saliency maps of intermediate layers, highly activated neurons always have the lion's share of relevance.

Fig.~\ref{fig:2} illustrates the motivational examples of trade-off among attribution methods (the same images in Fig.~\ref{fig:intro}) and presents the result of Grad-CAM, LRP, CEP, and RAP, which are based on a modified backpropagation algorithm with their individual purposes. As is widely known, Grad-CAM shows impressive localization performance for finding attributions of DNN predictions. However, because it utilizes the end layer of the feature extraction stage and interpolation, there are many losses in describing the details of pixel-level granularity.

The other methods in Fig.~\ref{fig:2} could fully decompose DNNs in a backward manner, including the feature extraction stage. The attributions from Integrated gradient provide the responsibility for a target label by computing the gradient w.r.t the features of the image. However, it is difficult to intuitively judge the quality of interpretation in a human-view due to the scattered and overlapped positive and negative attributions. Guided Backprop and LRP represent the output logit as the relevance scores in a pixel-level. However, although there is a minor difference in values, there is no visual difference in attributions between classes. The role of the CNN feature extractor stage goes from low-level features (edge or color) to high-level features (object or texture) as the layer becomes deeper. It is inevitable that the low-level features activated in the front stage, which turns out to be irrelevant in the latter part, are assigned the positive relevance in the backward layer-wise propagation. In conjunction with the above problem, unrelated parts to the target class, e.g. corner and watermark, tend to be attributed as positive.

While RAP approaches with the influential perspective to separate relevant/irrelevant attributions and shows strong advantages of objectness, it is not distinguishable among the predicted classes. As a countermeasure, \cite{zhang2018top} proposed a contrastive perspective, which contrasts the relevance for one class with the ones of all the others. However, the positive or negative relevance is distributed in the background or other parts not related to its origin. As this method shows a relatively activated area compared to other classes, there is a probability of having positive or negative relevance scores in irrelevant parts.

In this paper, the concept of hostile refers to an element that could have a negative influence on finding the attribution corresponding to the target. For example, when we decompose the output of the right image in Fig.~\ref{fig:intro} from the bottle class, the hostile class would be a person. The relevance of the hostile class is represented as hostile attributions. By assigning the negative relevance scores to the hostile attributions, we thwart the characteristics of ``winner always wins'', resulting in assigning bi-polar relevance scores among the target (positive), hostile (negative) attributions. We also present our method with a contrastive perspective for the hostile class. In this case, all other classes except the target are set hostile.

\section{Relative Sectional Propagation}
Inspired by the above problems, our method has two main streams: (i) Relative Gradient Activation Map and purging process, and (ii) Sectional propagation according to influence with gradient and uniform shifting.

\subsubsection{Relative Gradient Activation Map}
Letter \(y\) denotes the value of the network output before passing through the final layer of the classification stage. $\{t, o_1,\dots o_n, b\}$ is class notation where each notation represents a target, other predictions, and irrelevant classes, respectively. First, we obtain the gradient activation map $G$ by backpropagating the gradient of \(y\) until the end convolution layer $X$ of the feature extraction stage.
\begin{align}
    \begin{gathered}
        G^{(t)}_{ijk} = \lambda * ReLu\left(x_{ijk} * \frac{1}{Z} \sum_{i}\sum_{j}\frac{\partial y^t}{\partial x_{ijk}}\right)\\
        F_{ijk}^{(t)} = n*G^{(t)}_{ijk} - \sum_{q=1}^n G^{(h_q)}_{ijk}
    \end{gathered}
\end{align}
Here, $x_{ijk}$ denotes the neuron in the $k^{th}$ feature map of layer $X$, indexed by width $i$ and height $j$ dimensions, and $\lambda$ is a normalization factor to keep a maximum value as $1$. The equation for computing $G_{ijk}^{(t)}$ is the same process in Grad-CAM except for the last linear combination between the feature map $X$ and partial linearization. $h$ represents the hostile classes $\{o_1, \dots, o_n\}$.

The comparative gradients of the target against hostile class are contained in $F_{ijk}^{(t)}$. The attributions that are conflict to the channel attribution map ~\cite{fong2019understanding}, generated by summing over the channel dimension $k$, still exist along the channel. Conflicting attributions refer to the units in $F_{ijk}^{(t)}$ with opposite sign to the channel attribution. To preserve the gap between target and hostile attributions during propagation, it is necessary to eliminate these conflicting attributions as follows.
\begin{equation}
F^{\prime(t)}_{ijk} = 
\begin{cases}
 F_{ijk}^{(t)}  &,    \text{sign}\left(F_{ijk}^{(t)}\right) = \text{sign}\left(\sum_{k}F_{ijk}^{(t)}\right) \\
0 &,    \text{sign}\left(F_{ijk}^{(t)}\right)  \neq \text{sign}\left(\sum_{k}F_{ijk}^{(t)}\right) 
\end{cases}
\end{equation}
\begin{figure}
  \centering
  \includegraphics[width=1\linewidth]{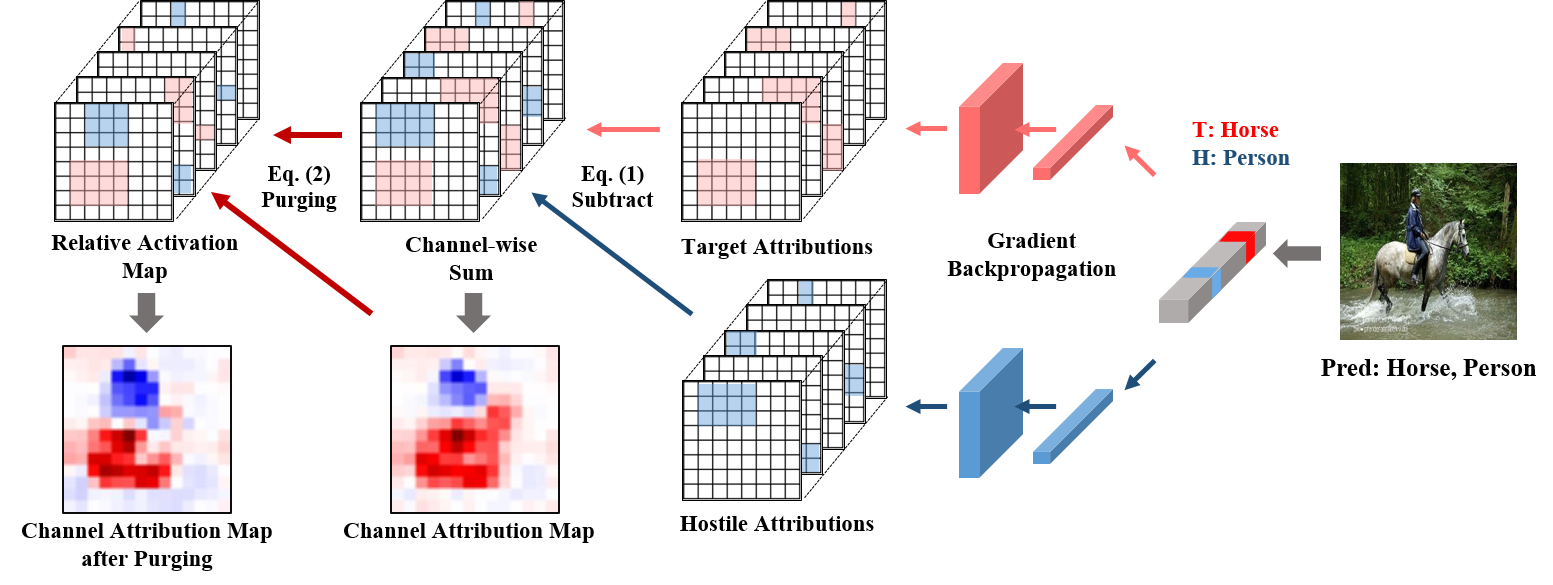}
    \caption{An illustration of generating the relative gradient activation map. The elements marked with red and blue color represent the target: \emph{Horse} and hostile: \emph{Person} attributions, respectively. Eq.~(1) computes the comparative gradients of the target. Eq.~(2) is the purging process to eliminate the conflicting attributions to the channel attribution map.}
  \label{fig:purge}
\end{figure}

\begin{figure}[t!]
  \centering
  \includegraphics[width=1\linewidth]{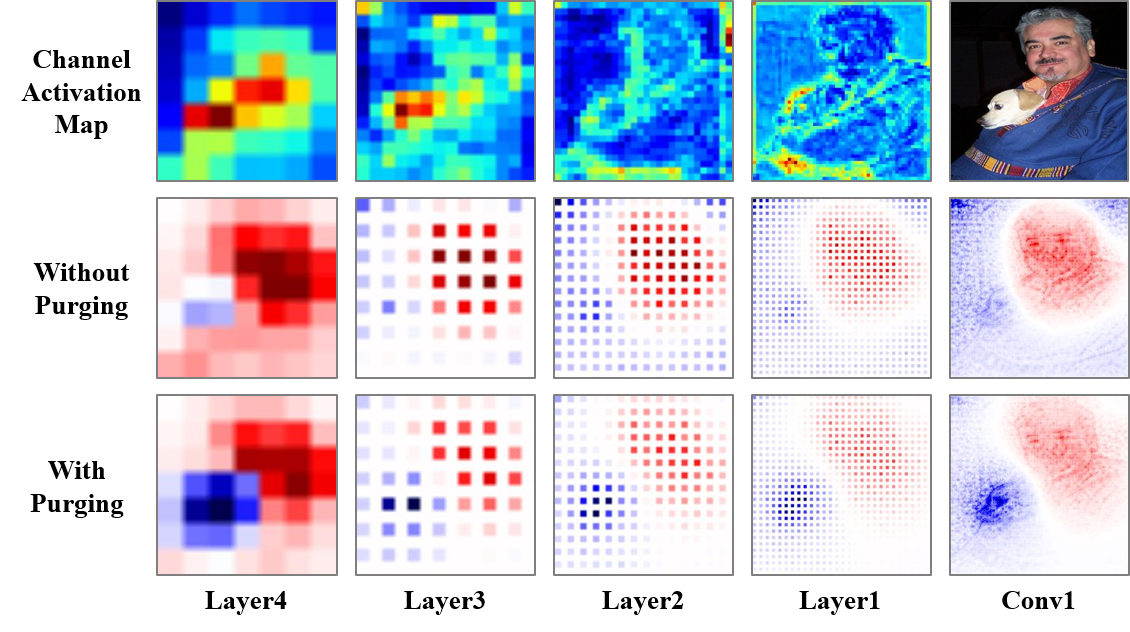}
    \caption{The difference between the channel attributions of intermediate layers with/without the purging process.}
  \label{fig:inter}
\end{figure}

Fig.~\ref{fig:purge} and Fig.~\ref{fig:inter} show the overview of generating the relative gradient activation map and the effect of the purging process to attributions in the intermediate layer of ResNet-50, respectively. In Fig.~\ref{fig:inter}, the first row illustrates the channel activation maps of intermediate layers, and most activations are concentrated in the dog region. After applying Eq.~(1) in case \{$t$=person, $h$=dog\}, the channel attribution map is shown as the first column. Without the purging process, although the dog regions are negative in the channel attribution map, there are still conflicting units (positive) along the channel. When we backpropagate each attribution in this state, it is inevitable to have an adverse effect on the backward step due to the nature of ``winner always wins''. Thus, they are canceled out and the gap between bi-polar relevance is decreased during the propagation procedure. As shown in the second row, instead of exact positions of hostile attributions, irrelevant parts, such as corners, are emphasized as negative. The phenomenon that the attribution is visualized as tiles is due to skip-connection operations in ResNet.

After, we have non-overlapping positive and negative attributions along the channel in $F^{\prime(t)}_{ijk}$ through channel-wise purging. Here, we set the positive and negative sections as \(\mathcal{P}_{ijk}^{(t)} = \{i,j,k| F_{ijk}^{\prime(t)}>0\}\) and \(\mathcal{N}_{ijk}^{(t)} = \{i,j,k| F_{ijk}^{\prime(t)}<0\}\). We normalize these values to make the sum of positive values have twice as many values as the sum of negative values: $\mathcal{P}_{ijk}^{(t)} \leftarrow 2*\mathcal{P}_{ijk}^{(t)}$, $R^{(t)}_{ijk} = \mathcal{P}_{ijk}^{(t)} \cup \mathcal{N}_{ijk}^{(t)}$. This normalization is necessary to preserve the conservation rule and prevent degeneracy problems. 

We report the difference between the two cases of obtaining the relative activation map $R^{(t)}_{ijk}$, (i) from Eq.~(1), and (ii) applying the contrastive perspective instead of Eq.~(1). For the latter case, we obtain $F^{(t)}_{ijk}$ by applying the contrastive excitation backprop \cite{zhang2018top} until the same convolution layer $X$ without channel-wise sum and ReLu. This is conceptually the same in terms of computing the relative gradient map for target class with the ones of all other (hostile) classes. The purging process and normalization are equally applied. Since the configuration and working mechanism are different according to the type of network, we report the results of both perspectives. The detailed equation of the latter case is described in the supplementary material.
\subsubsection{Sectional Relevance Propagation with Gradient}
The forward process between the current layer $l+1$ and the layer $l$ to which attributions are propagated is denoted as $f(x,w^{(l,l+1)})$. Also, we notate the boolean masks of \(\mathcal{P}_{ijk\in(l+1)}^{(t)}\) and \(\mathcal{N}_{ijk\in(l+1)}^{(t)}\) as $\mathcal{B}_{ijk\in(l+1)}^{+(t)}$, $\mathcal{B}_{ijk\in(l+1)}^{-(t)}$, respectively. Since $w^{(l,l+1)}$ are not directly influenced to $R^{(t)}_{ijk\in(l+1)}$, it is necessary to compute the gradient between $R^{(t)}_{ijk\in(l+1)}$ and the sectional influence of individual neurons $f(x,w^{(l,l+1)})*\mathcal{B}_{ijk\in(l+1)}^{\pm(t)}$ with respect to weight $w^{(l,l+1)}$. This gradient contains the correlation between the individual contributions of each neuron in a forward pass and the attributions in $R^{(t)}_{ijk\in(l+1)}$.
\begin{equation}
\begin{gathered}
    \nu^{+} = \frac{f(x,w^{(l,l+1)}) * \mathcal{B}_{ijk\in(l+1)}^{+(t)}}{\partial w^{(l,l+1)}} \mathcal{P}_{ijk\in(l+1)}^{(t)}\\
    \nu^{-} = \frac{f(x,w^{(l,l+1)}) * \mathcal{B}_{ijk\in(l+1)}^{-(t)}}{\partial w^{(l,l+1)}} \mathcal{N}_{ijk\in(l+1)}^{(t)}
\end{gathered}
\end{equation}
Here, Eq.~(3) could be represented as easily readable format: $\nu=f^*(f(x,w)*B, w, R)$ (vector-Jacobian product), which is implemented in many deep learning libraries and highly optimized. Through this gradient $\nu$, we backpropagate the attributions in $R^{(t)}_{ijk\in(l+1)}$ to the previous layer $l$ with an influential perspective \cite{nam2019relative} of individual neurons. 
\begin{equation}
\begin{split}
\hat R_{ijk\in(l)}^{\{t\}} & = x\odot f^{*}(f(x,\nu^{+}), \nu^{+}, \mathcal{P}_{ijk\in(l+1)}^{(t)} \oslash f(x, \nu^{+}))\\
& + x\odot f^{*}(f(x,\nu^{+}), \nu^{-}, \mathcal{N}_{ijk\in(l+1)}^{(t)} \oslash f(x, \nu^{-}))
\end{split}
\end{equation}

\begin{figure*}[t!]
  \centering
  \includegraphics[width=1\linewidth]{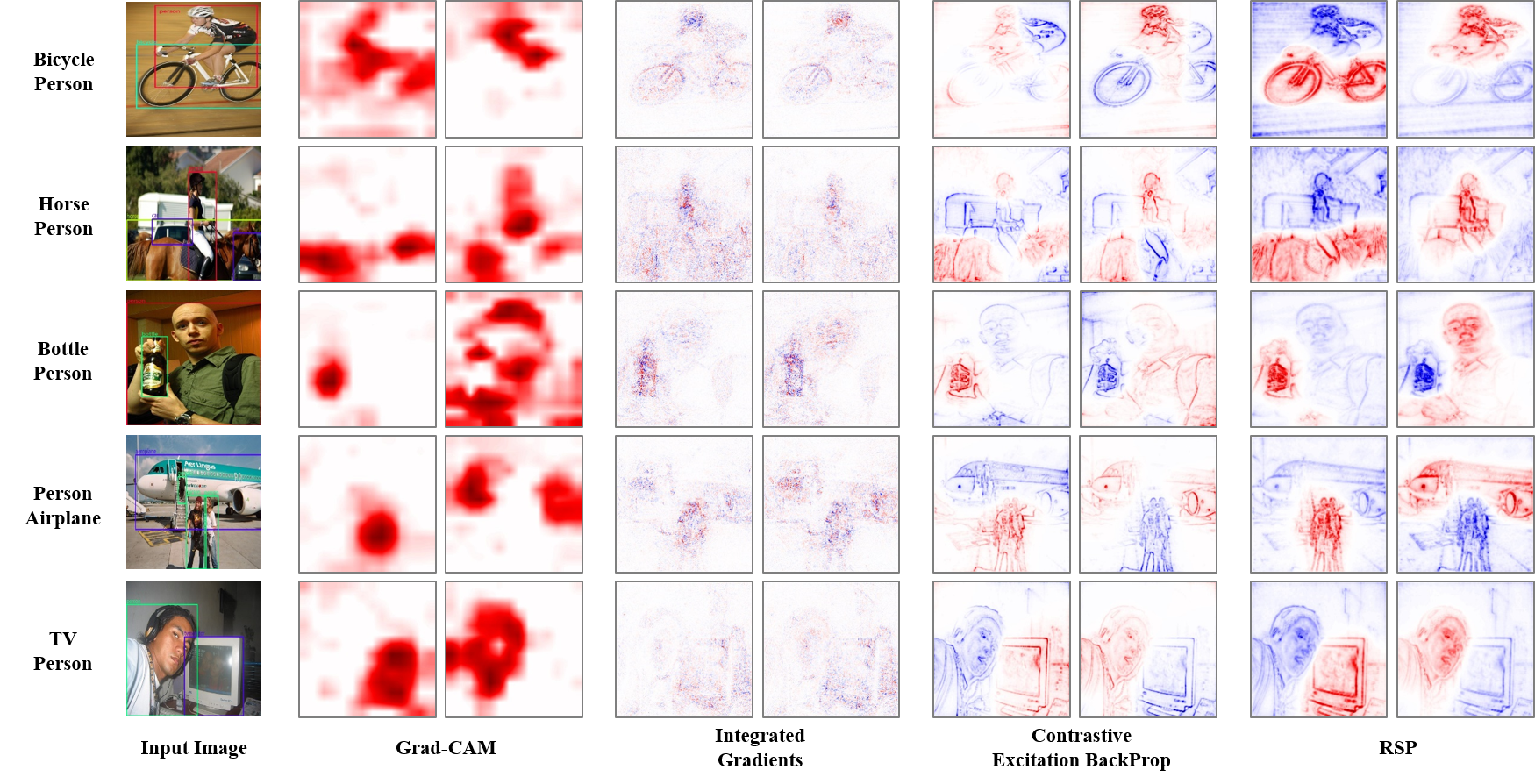}
    \caption{Comparison of the conventional attribution methods and RSP applied to VGG-16. The class names on the left side represent the predictions of DNNs among labels in input image. Each method shows the attribution results per predicted class.}
  \label{fig:all}
\end{figure*}

Here, $\odot$ and $\oslash$ denote the element-wise multiplication and division, respectively. The influence perspective represents that the importance of each neuron is an order of the absolute value, not the sign of the value. To be more specific, the attributions of the highest (positive) and lowest (negative) relevance scores denote a large amount of influence on the target and hostile classes, respectively. Attributions with a relevance score near-zero mean having a relatively small influence. From this influential perspective, it is possible to assign the relevance scores to activated neurons in order of importance among intermediate layers.

Now, the whole relevance sum of $\hat R_{ijk\in(l)}^{(t)}$ is the same as the sum of $R^{(t)}_{ijk\in(l+1)}$. We utilize uniform shifting ~\cite{nam2019relative} to change the irrelevant attributions, in which relevance scores are near zero, into negative. Let \(\Gamma\) be the number of activated neurons in $l$ and the sum of $\hat R_{ijk\in(l)}^{(t)}$ is $S$, this sum value is evenly divided and subtracted. To preserve the entire relevance sum as $S$ for each layer, we double the values of attributions in $\hat R_{ijk\in(l)}^{(t)}$.
\begin{equation}
\ddot R_{ijk\in(l)}^{(t)} = 
\begin{cases}
2*\hat R_{ijk\in(l)}^{(t)} - S * \frac{1}{\Gamma}  &,    \text{ $x_{ijk}>0$ } \\
2*\hat R_{ijk\in(l)}^{(t)} &,    \text{ $x_{ijk}=0$ } 
\end{cases}
\end{equation}
In case that a neuron $x_{ijk}$ is not activated, $\hat R_{ijk\in(l)}^{(t)}$ is equal to zero. Relatively unimportant attributions, which are near zero, would be converted into the negative during the propagation procedure, thereby the irrelevant attributions, e.g. background, have the negative relevance scores in the final output.
$\ddot R_{ijk\in(l)}^{(t)}$ is the input attributions for the next previous layer $l-1$ and propagated by repeating this process from the purging step. This procedure is repeated until the first layer $l=1$ of the model. For the final propagation between the input and first layer, we adopt the \(Z^{\beta}\) rule \cite{bach2015pixel} which is commonly used for propagating to the input layer, resulting in clear visualizations without distorting the priority of attributions. Detail expansion of each equation is described in the supplementary material.

\begin{table*}[t]
\RawFloats
  \centering
    \resizebox{1\columnwidth}{!}{
    \large
    \begin{tabular}{cc|cc|cc|cc|cc|cc|cc|cc|cc}
        \specialrule{2.0pt}{1pt}{1pt}
     & &  \multicolumn{8}{c|}{\textbf{PASCAL VOC 2007}} & \multicolumn{8}{c}{\textbf{COCO 2014}} \bigstrut[b]\\
          & &\multicolumn{4}{c|}{VGG-16}  &\multicolumn{4}{c|}{ResNet-50} &\multicolumn{4}{c|}{VGG-16}  &\multicolumn{4}{c}{ResNet-50}  \bigstrut[b]\\
          & &\multicolumn{2}{c}{ALL}&\multicolumn{2}{c|}{DIF}  &\multicolumn{2}{c}{ALL}&\multicolumn{2}{c|}{DIF} &\multicolumn{2}{c}{ALL}&\multicolumn{2}{c|}{DIF}  &\multicolumn{2}{c}{ALL}&\multicolumn{2}{c}{DIF}  \bigstrut[b]\\
    METHOD &T& PG & mIOU  & PG & mIOU & PG & mIOU  & PG & mIOU  & PG & mIOU  & PG & mIOU  & PG & mIOU  & PG & mIOU\\
    \hline
    \multirow{2}[2]{*}{Grad-CAM} & L & .866  & .43/.49 
    & .740  & .39/.48
    & .903  & .56/.57 
    & .823  & .47/.57 
    & .542  & .35/.46 
    & .490  & .33/.43  
    & .573  & .44/.51  
    & .523  & .40/.48      \bigstrut[t]\\
          
    & P & .945  & .41/.50 
    & .924  & .33/.54 
    & .953  & .55/.58 
    & .932  & .44/.59 
    & .727  & .30/.49 
    & .689  & .25/.45 
    & .705  & .39/.52 
    & .674  & .32/.47 \bigstrut[b]\\
    \hdashline
    \hdashline
    \multirow{2}[2]{*}{Gradient} & L & .762  & \textcolor[gray]{0.5}{.00}/.47 
    & .568  & \textcolor[gray]{0.5}{.00}/.41 
    & .723  & \textcolor[gray]{0.5}{.00}/.45 
    & .568  & \textcolor[gray]{0.5}{.00}/.40 
    & .355  & \textcolor[gray]{0.5}{.00}/.39 
    & .289  & \textcolor[gray]{0.5}{.00}/.37 
    & .319  & \textcolor[gray]{0.5}{.00}/.39
    & .262  & \textcolor[gray]{0.5}{.00}/.37      \bigstrut[t]\\
          
    & P&  .858  & \textcolor[gray]{0.5}{.00}/.49 
    & .716  & \textcolor[gray]{0.5}{.00}/.50 
    & .734  & \textcolor[gray]{0.5}{.00}/.44
    & .605  & \textcolor[gray]{0.5}{.00}/.43
    & .547  & \textcolor[gray]{0.5}{.00}/.44
    & .492  & \textcolor[gray]{0.5}{.00}/.40
    & .455  & \textcolor[gray]{0.5}{.00}/.42
    & .405  & \textcolor[gray]{0.5}{.00}/.38  \bigstrut[b]\\
    \hdashline
    \multirow{2}[2]{*}{DeconvNet} & L & .675 & \textcolor[gray]{0.5}{.00}/.41 
    & .441  & \textcolor[gray]{0.5}{.00}/.31 
    & .686  & \textcolor[gray]{0.5}{.00}/.43 
    & .447  & \textcolor[gray]{0.5}{.00}/.33 
    & .241  & \textcolor[gray]{0.5}{.00}/.35  
    & .164  & \textcolor[gray]{0.5}{.00}/.32 
    & .273  & \textcolor[gray]{0.5}{.00}/.35 
    & .192  & \textcolor[gray]{0.5}{.00}/.33       \bigstrut[t]\\
          
    & P  & .802  & \textcolor[gray]{0.5}{.00}/.46
    & .573  & \textcolor[gray]{0.5}{.00}/.37 
    & .789  & \textcolor[gray]{0.5}{.00}/.44
    & .595  & \textcolor[gray]{0.5}{.00}/.39 
    & .469  & \textcolor[gray]{0.5}{.00}/.36 
    & .372  & \textcolor[gray]{0.5}{.00}/.31 
    & .429  & \textcolor[gray]{0.5}{.00}/.36 
    & .338  & \textcolor[gray]{0.5}{.00}/.31  \bigstrut[b]\\
    \hdashline
    \multirow{1}[2]{*}{Guided} & L & .758 & \textcolor[gray]{0.5}{.00}/\textcolor[rgb]{ .0,0,1}{\textbf{.49}}  
    & .530  & \textcolor[gray]{0.5}{.00}/.43  
    & .771  & \textcolor[gray]{0.5}{.00}/.51  
    & .594  & \textcolor[gray]{0.5}{.00}/.46  
    & .365  & \textcolor[gray]{0.5}{.00}/.41  
    & .288  & \textcolor[gray]{0.5}{.00}/.39
    & .410  & \textcolor[gray]{0.5}{.00}/.43  
    & .340  & \textcolor[gray]{0.5}{.00}/.41       \bigstrut[t]\\
          
    \multirow{1}[2]{*}{BackProp}& P & \textcolor[rgb]{ .0,0,0}{{.880}} & \textcolor[gray]{0.5}{.00}/\textcolor[rgb]{ .753,0,0}{\textbf{.52}}  
    & .784  & \textcolor[gray]{0.5}{.00}/.54  
    & .857  & \textcolor[gray]{0.5}{.00}/.53 
    & .756  & \textcolor[gray]{0.5}{.00}/.53 
    & .600  & \textcolor[gray]{0.5}{.00}/.47
    & .536  & \textcolor[gray]{0.5}{.00}/.43 
    & .573  & \textcolor[gray]{0.5}{.00}/.47 
    & .519  & \textcolor[gray]{0.5}{.00}/.44 \bigstrut[b]\\
    \hdashline

    \multirow{1}[2]{*}{Excitation} & L & .735 & \textcolor[gray]{0.5}{.00}/.46
    & .520  & \textcolor[gray]{0.5}{.00}/.45 
    & .785  & \textcolor[gray]{0.5}{.00}/.46
    & .623  & \textcolor[gray]{0.5}{.00}/.45
    & .377  & \textcolor[gray]{0.5}{.00}/.42
    & .304  & \textcolor[gray]{0.5}{.00}/.40
    & .437  & \textcolor[gray]{0.5}{.00}/.43 
    & .374  & \textcolor[gray]{0.5}{.00}/.41      \bigstrut[t]\\
          
    \multirow{1}[2]{*}{BackProp} & P & .856 & \textcolor[gray]{0.5}{.00}/.47
    & .742  & \textcolor[gray]{0.5}{.00}/.53 
    & .864  & \textcolor[gray]{0.5}{.00}/.47
    & .768  & \textcolor[gray]{0.5}{.00}/.50
    & .573  & \textcolor[gray]{0.5}{.00}/.47 
    & .505  & \textcolor[gray]{0.5}{.00}/.45
    & .582  & \textcolor[gray]{0.5}{.00}/.46
    & .533  & \textcolor[gray]{0.5}{.00}/.44 \bigstrut[b]\\
    \hdashline
    \multirow{1}[2]{*}{c*Exitation} & L & .766 & .38/.45 
    & \textcolor[rgb]{ .0,0,1}{\textbf{.634}}  & .34/.50 
    & .857  & .49/.49 
    & .741  & \textcolor[rgb]{ .0,0,1}{\textbf{.45}} /\textcolor[rgb]{ .753,0,0}{\textbf{.56}} 
    & \textcolor[rgb]{ .0,0,0}{{.472}}  & .32/.46 
    & \textcolor[rgb]{ .0,0,0}{{.417}}  & .30/.45 
    & .536  & \textcolor[rgb]{.753,0,0}{\textbf{.41}}/\textcolor[rgb]{.753,0,0}{\textbf{.49}}
    & .485  & \textcolor[rgb]{.753,0,0}{\textbf{.37}}/\textcolor[rgb]{.753,0,0}{\textbf{.48}}      \bigstrut[t]\\
          
    \multirow{1}[2]{*}{BackProp}& P & .856 & .40/.42 
    & \textcolor[rgb]{ .0,0,0}{{.784}}  & .39/.55
    & \textcolor[rgb]{ .0,0,1}{\textbf{.945}}  & .52/.49 
    & \textcolor[rgb]{ .0,0,1}{\textbf{.887}}  & \textcolor[rgb]{ .0,0,1}{\textbf{.51}}/\textcolor[rgb]{ .753,0,0}{\textbf{.62}} 
    & \textcolor[rgb]{ .0,0,0}{{.659}}  & .37/.49 
    & \textcolor[rgb]{ .0,0,0}{{.620}}  & .34/.50 
    & .671  & \textcolor[rgb]{.0,0,1}{\textbf{.47}}/\textcolor[rgb]{.753,0,0}{\textbf{.53}} 
    & .636  & \textcolor[rgb]{.0,0,1}{\textbf{.42}}/\textcolor[rgb]{.753,0,0}{\textbf{.53}} \bigstrut[b]\\
    \hdashline
    \rowcolor[rgb]{ .906,  .902,  .902}     \multirow{2}[2]{*}{RSP}  & L
    & \textcolor[rgb]{ .753,0,0}{\textbf{.849}}
    & \textcolor[rgb]{ .753,0,0}{\textbf{.51}}/\textcolor[rgb]{ .753,0,0}{\textbf{.51}}
    & \textcolor[rgb]{ .753,0,0}{\textbf{.712}}
    & \textcolor[rgb]{ .753,0,0}{\textbf{.43}}/\textcolor[rgb]{ .753,0,0}{\textbf{.54}}
    
    & \textcolor[rgb]{ .0  ,0,1}{\textbf{.859}}
    & \textcolor[rgb]{ .0,0,1}{\textbf{.49}}/\textcolor[rgb]{ .0,0,1}{\textbf{.51}}
    & \textcolor[rgb]{ .0,0,1}{\textbf{.749}}
    & \textcolor[rgb]{ .0,0,0}{.39}/\textcolor[rgb]{ .0,0,0}{.49}
    
    & \textcolor[rgb]{ .753,0,0}{\textbf{.540}}
    & \textcolor[rgb]{ .753,0,0}{\textbf{.43}}/\textcolor[rgb]{ .753,0,0}{\textbf{.49}}
    & \textcolor[rgb]{ .753,0,0}{\textbf{.479}}
    & \textcolor[rgb]{ .753,0,0}{\textbf{.37}}/\textcolor[rgb]{ .753,0,0}{\textbf{.47}}
    
    & \textcolor[rgb]{ .753,0,0}{\textbf{.558}}
    & \textcolor[rgb]{ .0,0,0}{{.39}}/\textcolor[rgb]{ .0,0,0}{{.46}}
    & \textcolor[rgb]{ .753,0,0}{\textbf{.504}}
    & \textcolor[rgb]{ .0,0,0}{{.35}}/\textcolor[rgb]{ .0,0,0}{{.43}}
    \bigstrut[t]\\
    
    \rowcolor[rgb]{ .906,  .902,  .902} 
     \multirow{-2}{*}{RSP} & P
    & \textcolor[rgb]{ .753,0,0}{\textbf{.946}}
    & \textcolor[rgb]{ .753,0,0}{\textbf{.56}}/\textcolor[rgb]{ .0,0,1}{\textbf{.51}}
    & \textcolor[rgb]{ .753,0,0}{\textbf{.903}}
    & \textcolor[rgb]{ .753,0,0}{\textbf{.54}}/\textcolor[rgb]{ .753,0,0}{\textbf{.63}}
    
    & \textcolor[rgb]{ .0,0,0}{.909}
    & \textcolor[rgb]{ .0,0,1}{\textbf{.54}}/\textcolor[rgb]{ .0,0,1}{\textbf{53}}
    & \textcolor[rgb]{ .0,0,0}{.836}
    & \textcolor[rgb]{ .0,0,0}{.44}/\textcolor[rgb]{ .0,0,0}{54}
    
    & \textcolor[rgb]{ .753,0,0}{\textbf{.725}}
    & \textcolor[rgb]{ .753,0,0}{\textbf{.51}}/\textcolor[rgb]{ .753,0,0}{\textbf{56}}
    & \textcolor[rgb]{ .753,0,0}{\textbf{.680}}
    & \textcolor[rgb]{ .753,0,0}{\textbf{.45}}/\textcolor[rgb]{ .753,0,0}{\textbf{54}}
    
    & \textcolor[rgb]{ .0,0,1}{\textbf{.688}}
    & \textcolor[rgb]{ .0,0,0}{{.44}}/\textcolor[rgb]{ .0,0,0}{{.51}}
    & \textcolor[rgb]{ .0,0,1}{\textbf{.654}}
    & \textcolor[rgb]{ .0,0,0}{{.38}}/\textcolor[rgb]{ .0,0,0}{{.48}}
     \bigstrut[t] \\
     \hdashline
    \rowcolor[rgb]{ .906,  .902,  .902}     \multirow{2}[2]{*}{c*RSP}  & L
    & \textcolor[rgb]{ .0,0,1}{\textbf{.785}}
    & \textcolor[rgb]{ .0,0,1}{\textbf{.46}}/\textcolor[rgb]{ .0,0,0}{.47}
    & \textcolor[rgb]{ .0,0,0}{{.627}}
    & \textcolor[rgb]{ .0,0,1}{\textbf{.42}}/\textcolor[rgb]{ .0,0,1}{\textbf{.52}}
    
    & \textcolor[rgb]{ .753,0,0}{\textbf{.891}}
    & \textcolor[rgb]{ .753,0,0}{\textbf{.52}}/\textcolor[rgb]{ .753,0,0}{\textbf{.52}}
    & \textcolor[rgb]{ .753,0,0}{\textbf{.777}}
    & \textcolor[rgb]{ .753,0,0}{\textbf{.46}}/\textcolor[rgb]{ .0,0,1}{\textbf{.54}}
    
    & \textcolor[rgb]{ .0,0,1}{\textbf{.475}}
    & \textcolor[rgb]{ .0,0,1}{\textbf{.39}}/\textcolor[rgb]{ .0,0,1}{\textbf{.47}}
    & \textcolor[rgb]{ .0,0,1}{\textbf{.418}}
    & \textcolor[rgb]{ .0,0,1}{\textbf{.36}}/\textcolor[rgb]{ .0,0,1}{\textbf{.45}}
    
    & \textcolor[rgb]{ .0,0,1}{\textbf{.545}}
    & \textcolor[rgb]{ .0,0,1}{\textbf{.41}}/\textcolor[rgb]{ .0,0,1}{\textbf{.47}}
    & \textcolor[rgb]{ .0,0,1}{\textbf{.488}}
    & \textcolor[rgb]{ .0,0,1}{\textbf{.37}}/\textcolor[rgb]{ .0,0,1}{\textbf{.44}}
    \bigstrut[t]\\
    
    \rowcolor[rgb]{ .906,  .902,  .902} 
     \multirow{-2}{*}{c*RSP} & P
    & \textcolor[rgb]{ .0,0,1}{\textbf{.881}}
    & \textcolor[rgb]{ .0,0,1}{\textbf{.49}}/\textcolor[rgb]{ .0,0,0}{.46}
    & \textcolor[rgb]{ .0,0,1}{\textbf{.791}}
    & \textcolor[rgb]{ .0,0,1}{\textbf{.51}}/\textcolor[rgb]{ .0,0,1}{\textbf{.60}}
    
    & \textcolor[rgb]{ .753,0,0}{\textbf{.949}}
    & \textcolor[rgb]{ .753,0,0}{\textbf{.56}}/\textcolor[rgb]{ .753,0,0}{\textbf{.53}}
    & \textcolor[rgb]{ .753,0,0}{\textbf{.893}}
    & \textcolor[rgb]{ .753,0,0}{\textbf{.53}}/\textcolor[rgb]{ .0,0,1}{\textbf{.61}}
    
    & \textcolor[rgb]{ .0,0,1}{\textbf{.675}}
    & \textcolor[rgb]{ .0,0,1}{\textbf{.46}}/\textcolor[rgb]{ .0,0,1}{\textbf{.51}}
    & \textcolor[rgb]{ .0,0,1}{\textbf{.634}}
    & \textcolor[rgb]{ .0,0,1}{\textbf{.42}}/\textcolor[rgb]{ .0,0,1}{\textbf{.52}}
    
    & \textcolor[rgb]{ .753,0,0}{\textbf{.697}}
    & \textcolor[rgb]{ .753,0,0}{\textbf{.47}}/\textcolor[rgb]{ .0,0,1}{\textbf{.52}}
    & \textcolor[rgb]{ .753,0,0}{\textbf{.659}}
    & \textcolor[rgb]{ .753,0,0}{\textbf{.42}}/\textcolor[rgb]{ .0,0,1}{\textbf{.49}}
     \bigstrut[t] \\
    \specialrule{2.0pt}{1pt}{1pt}
    \end{tabular}%
    }
    \caption{The performance of Pointing Game and mIOU over Pascal VOC 2007 test set and COCO 2014 validation set. $T$ denotes the different circumstance of each step for testing: $P$: Only predicted classes, $L$: All labels. ALL and DIF represent the full data and the subset of difficult images, respectively. For the mIOU results, left/right values means the performance when applying without/with the threshold. The threshold is set as the mean value of positive attributions. \textcolor{myred}{RED} and \textcolor{myblue}{BLUE} represent the most and second highest numbers excluding Grad-CAM, respectively. All attribution methods, except Grad-CAM, is fully decomposed from the output to the \textbf{first} layer of each network.} \label{table:tab1}
\end{table*}%
\begin{figure}[t!]
  \centering
  \includegraphics[width=1\linewidth]{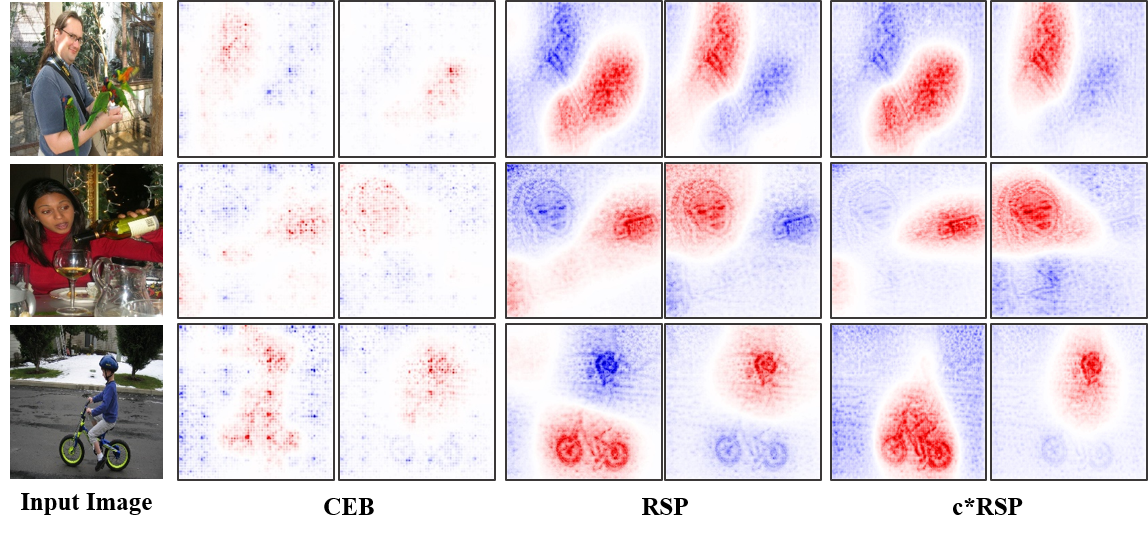}
    \caption{Comparison of CEB, RSP and RSP with contrastive perspective in ResNet-50.}
  \label{fig:res}
\end{figure}
\section{Experimental Evaluations}
\subsection{Implementation Details}
We utilize the popular CNN architectures: VGG-16 and ResNet-50. Each model is trained on Pascal VOC 2007 \cite{everingham2010pascal} and MS COCO 2014 dataset \cite{lin2014microsoft}, which are widely employed and easily accessible. For a fair comparison, the models that we used are available online with TorchRay package \cite{fong2019understanding}. We implement our method with PyTorch and visualize the attributions as a heatmap represented by seismic colors, where red and blue colors denote positive and negative values, respectively. We utilize the implementation introduced in \cite{fong2019understanding} for the other attribution methods. All conditions in experiments are the same except for setting the saliency layer. For the fully decomposed environment, the saliency layer for visualizing the attributions is set as the first convolution layer of each model. Each type of evaluation is described in subsequent sections.

\subsection{Qualitative Assessment}
To qualitatively evaluate the attributions of each method, we report the visual differences to compare how the high-rated points are gathered in the bounding box. As the goal of attribution methods are the same for seeking the most important factors, we could assess the consistency of positive relevance among methods. Among many researches related to attribution methods that are described in related work, we compare the methods that can be visualized with class-discriminativeness and contain detailed information of neuron activations. The other methods are not qualitatively compared due to the reasons related to Fig.~\ref{fig:2} (shown in supplementary material). The comparison methods are Grad-CAM, Integrated Gradients, Contrastive Excitation Backprop, and RSP. Fig.~\ref{fig:all} and Fig.~\ref{fig:res} illustrate the heatmaps of each method for the output predictions of VGG-16 and ResNet-50, respectively. Compared to other methods, RSP shows detailed descriptions of activated neurons and clear separations between the target object and other objects (including background), resulting in much clearer visualizations with strong objectness. More qualitative comparisons are given in the supplementary material.

\begin{figure}[h!]
  \centering
  \includegraphics[width=1\linewidth]{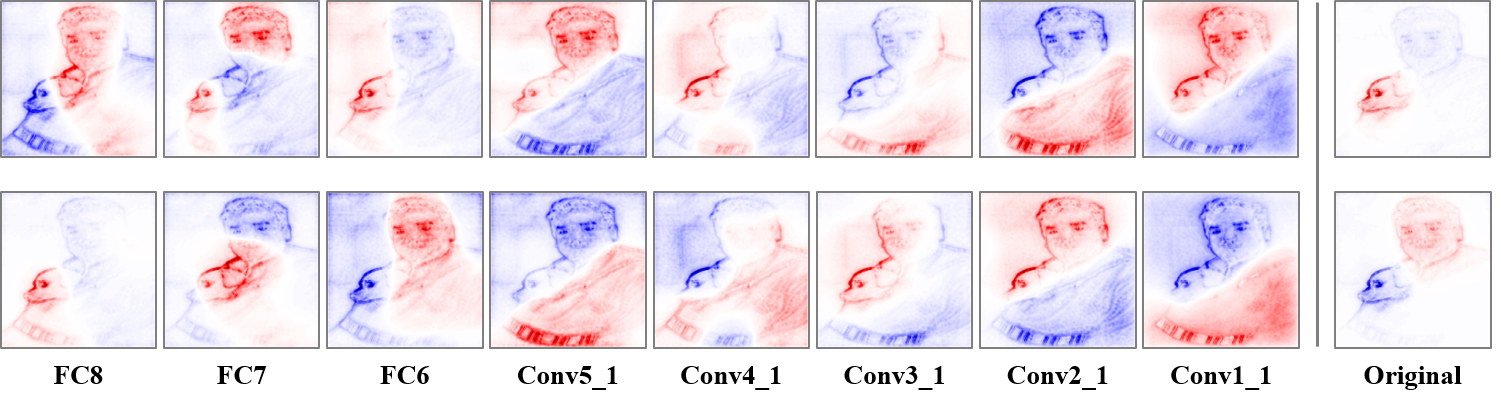}
    \caption{Model weights of VGG-16 are progressively initialized from the end to beginning.}
  \label{fig:sal}
\end{figure}
\subsubsection{Sanity Check}
\cite{adebayo2018sanity} addresses the non-sensitivity problem of some saliency methods when the parameters of the model are randomly initialized in a cascading fashion from the end layer. It is crucial to support the basis that our explanation and the model decision are dependent on each other. Fig.~\ref{fig:sal} illustrates the variations of our method when applying weight randomization progressively. Attributions from each label are extremely distorted compared to the original explanations. 
\subsection{Evaluating Quality of Attributions}
\begin{table}[t!]
\begin{small}
    \centering
    \begin{tabular}{@{}l|c@{}}
        \toprule[0.4ex]
        Method & mIOU\\
        \midrule
         Grad-CAM (threshold: mean) + CRF & 52.14\\
         Segmentation Prop~\cite{guillaumin2014imagenet} & 57.30\\
         DeepMask~\cite{pinheiro2015learning} &  58.69\\
         RAP~\cite{nam2019relative} &59.46\\
         DeepSaliency~\cite{li2016deepsaliency} &  62.12\\
         Pixel Objectness~\cite{xiong2018pixel} & 64.22\\
         
        \midrule
         RSP & 60.81\\
         \textcolor[rgb]{ .753,0,0}{\textbf{RSP + CRF}} & \textcolor[rgb]{ .753,0,0}{\textbf{64.51}}\\
         \bottomrule[0.4ex]
    \end{tabular}
    \end{small}
    \caption{Segmentation mIoU results on the ImageNet Segmentation task. Our method is highly comparable to those methods without using the additional supervision.}
    \label{imagenet_seg}
\end{table}
\subsubsection{Pointing Game}
Pointing Game \cite{zhang2018top} assesses the attribution methods by computing the matching scores of localization between the highest relevance point and the semantic annotations of object categories in the image. However, in previous literature, this metric does not consider the performance of the model itself because the assessed attributions are decomposed from the label, not the predicted classes. It is necessary to consider the context of the predictions because performing the decomposition on a label that has not been identified by DNNs is likely to lead to incorrect interpretation. Therefore, we report both cases of the decomposition from (i) P: only predicted labels and (ii) L: all labels in Tab.\ref{table:tab1}.

To compare the attribution methods in completely decomposed circumstances, all methods except Grad-CAM do backward propagation until the first convolution layer of each network. Methods with notation c* denote a contrastive perspective to set comparative classes to target. In our case, c* represents the case of computing the relative gradient activation map from the contrastive perspective. As shown in the results, our method shows the superior performance of localization in both cases compared to the fully decomposed attribution methods. In VGG-16, the relative gradient maps between the target and the rest of the predicted classes show much superior performance compared to the contrastive initialization. On the contrary, ResNet-50 shows much better results in the class setting of contrasting.  

\subsubsection{Objectness and Weakly Supervised Segmentation}
Interpretation of the attribution methods and objectness is closely related in terms of aiming to find the pixels corresponding to the target object. Based on this concept, many studies in weakly supervised segmentation (image-label level supervision) \cite{ahn2019weakly, Lee_2019_CVPR, huang2018weakly} start from the seeds extracted from the attribution methods. We report the mean Intersection of Union (mIoU) to measure the false positives of attributions that are distributed in irrelevant parts (other objects or background). Instead of comparing it with the segmentation mask, we computed the accuracy of mIOU between the bounding box and positive attributions to allow the error range in both datasets. Since some methods: Guided Backprop, DeconvNet, and Excitation Backprop, have all positive values in output results, the mean value of attributions is taken as the threshold. In Tab.\ref{table:tab1}, the left (right) values for each method represent the without (with) the threshold. Our method outperforms in almost all cases, especially when the target is in the set of correct predictions. 

\begin{figure}[t!]
  \centering
  \includegraphics[width=1\linewidth]{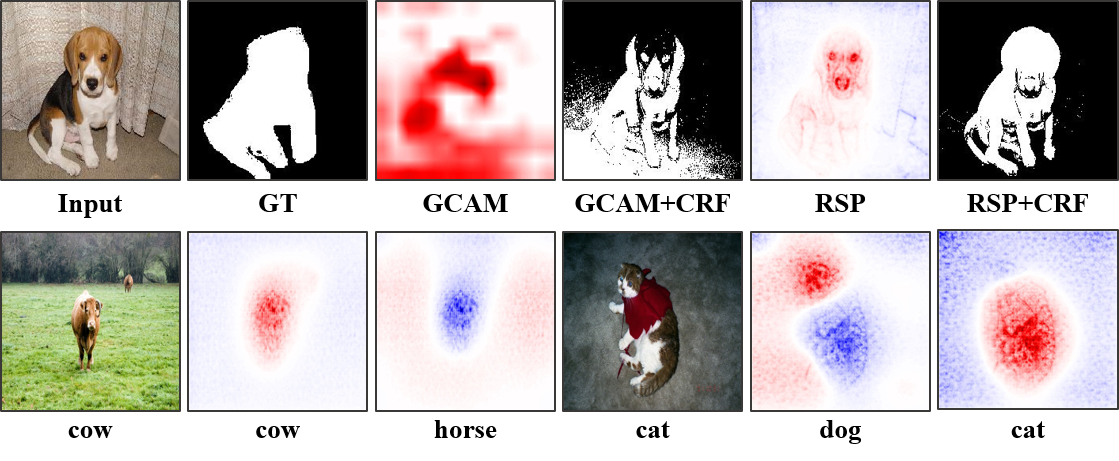}
    \caption{The first and second rows demonstrate the weakly supervised segmentation results with CRF and Misconception of ResNet-50 about a single object, respectively.}
  \label{fig:seg_dis}
\end{figure}

Furthermore, we represent the segmentation performance on the ImageNet segmentation dataset  ~\cite{Guillaumin2014ImageNetAW}, which consists of 4,276 images with segmentation mask. Tab.\ref{imagenet_seg} compares other saliency based objectness methods and RSP. The results of Grad-CAM, RAP, and RSP are from the ImageNet pretrained model in PyTorch. Although some methods use additional supervision, (i.e. bounding box, optical flow), RSP shows comparable performance with only image-label level supervision. First row in Fig.~\ref{fig:seg_dis} shows the results of Grad-CAM (mean threshold) and RSP with Dense-CRF \cite{krahenbuhl2011efficient} to refine the attributions, resulting in superior performance compared to other methods.
\section{Discussions}
To the best of our knowledge, there is still no exact elucidation of the internal mechanisms except for the structural and conceptual differences between VGG-Net and ResNet. Some differences could be inferred by investigating the failure attributions seen only in ResNet. ResNet tends to classify objects independently between classes. For example in Fig.~\ref{fig:seg_dis} second row, a single object is misclassified as two classes by focusing on different features, resulting in overlap of the relative gradient activation map from Eq.~(1). Furthermore, there is a clear effect of unlabeled objects in ResNet. In these cases, the contrastive hostile setting shows better performance to find the target attributions. More discussion and analysis are described in the supplementary materials. 
\section{Conclusion}
In this paper, we propose a new attribution method for decomposing the output of DNNs by assigning the bi-polar relevance score among the target and hostile classes. From the antagonistic perspective among objects, it is possible to allocate the bi-polar relevance scores to neuron activations, resulting in distinguishable and attentive decomposition. We assess our methods in quantitative and qualitative ways with i) Pointing Game, ii) mIOU, and iii) Model randomization to confirm the quality of attributions. The results demonstrate that the attributions from RSP have the properties of strong objectness, class-discriminativeness, and detailed descriptions of the neuron activations.

\section{Acknowledgments}
This work was supported by Institute for Information \& communications Technology Planning \& Evaluation (IITP) grant funded by the Korea government (MSIT) (No. 2017-0-01779, A machine learning and statistical inference framework for explainable artificial intelligence \& No. 2019-0-01371, Development of brain-inspired AI with human-like intelligence \& No. 2019-0-00079,  Artificial Intelligence Graduate School Program, Korea University).

{\small
\bibstyle{aaai21}
\bibliography{egbib}
}

\end{document}